\documentclass[10pt,twocolumn,letterpaper]{article}

\usepackage{times}
\usepackage{epsfig}
\usepackage{graphicx}
\usepackage{amsmath}
\usepackage{amssymb}
\usepackage{caption}
\usepackage{subcaption}
\usepackage{multirow} 

\newcommand{\Lagr}{\mathcal{L}}

\usepackage[pagebackref=true,breaklinks=true,letterpaper=true,colorlinks,bookmarks=false]{hyperref}

\begin{document}

\title{PLAN-B: Predicting Likely Alternative Next Best Sequences for Action Prediction}

\author{Dan Scarafoni\\
{\tt\small danscarafoni@gatech.edu}
\and
Irfan Essa
{\tt\small irfan@gatech.edu}

\and
Thomas Pl{\"o}tz
{\tt\small thomas.ploetz@gatech.edu}
}

\maketitle

\begin{abstract}
Action prediction focuses on anticipating actions before they happen. Recent works leverage probabilistic approaches to describe future uncertainties and sample future actions. However, these methods cannot easily find all alternative predictions, which are essential given the inherent unpredictability of the future, and current evaluation protocols do not measure a system's ability to find such alternatives. We re-examine action prediction in terms of its ability to predict not only the top predictions, but also top alternatives with the accuracy@k metric. In addition, we propose Choice F1: a metric inspired by F1 score which evaluates a prediction system's ability to find all plausible futures while keeping only the most probable ones. To evaluate this problem, we present a novel method, \textbf{P}redicting the \textbf{L}ikely \textbf{A}lternative \textbf{N}ext \textbf{B}est, or \textbf{PLAN-B}, for action prediction which automatically finds the set of most likely alternative futures. PLAN-B consists of two novel components: (i) a Choice Table which ensures that all possible futures are found, and (ii) a ``Collaborative'' RNN system which combines both action sequence and feature information. We demonstrate that our system outperforms state-of-the-art results on benchmark datasets.
\end{abstract}


\begin{figure*}
    \centering
    \includegraphics[height=15em,width=\linewidth]{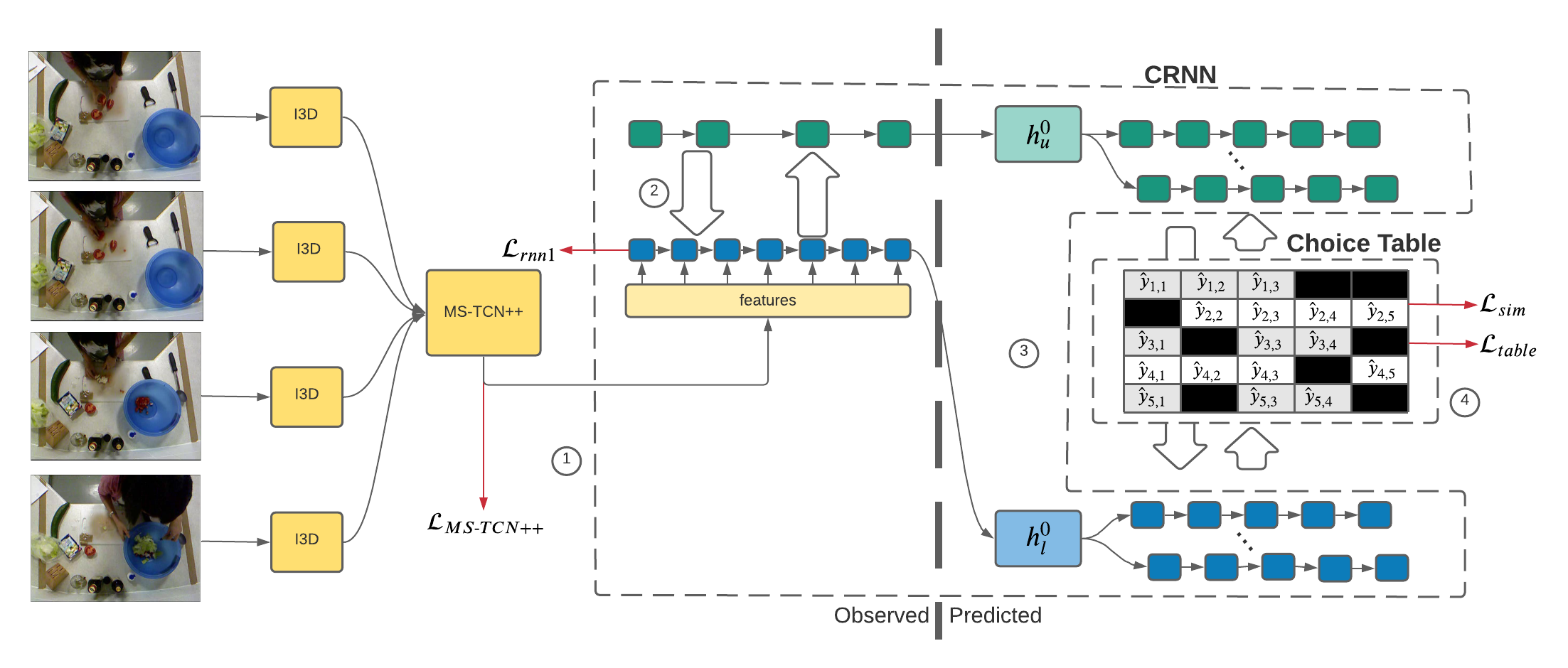}
    \caption{Overview of our system. (1) We encode video frame features with I3D and MS-TCN++ \cite{Carreira2017,li_ms-tcn_2020}. (2) The CRNN model forms a higher and lower level encoding of the action sequence information. (3) The decoders of the CRNN output predict action sequences and times (not shown) into the Choice Table. (4) During training, the Choice Table uses Similarity Penalty and Random Loss Negation to ensure all likely sequences are predicted.}
    \label{fig:main_system}
\end{figure*}

\vspace*{-1em}
\section{Introduction}
Action prediction aims to understand and anticipate future actions before they occur. Understanding the future is vital for a broad spectrum of problems. 
Robots collaborating with humans, for example, must be able to anticipate actions in order to safely work with their partners \cite{albrecht_autonomous_2018,heard_diagnostic_2018}. 
It is also essential in surveillance applications, e.g., for early detection of abnormal events \cite{ke_time-conditioned_2019}.

Recent efforts focus on predicting future video actions (action forecasting) attempting to predict future activity sequences (on the order of minutes) after observing a small portion of a video \cite{Farha2018b}. 
Most work in this area predicts either the next action or next series of actions with maximum accuracy \cite{Farha2018b,furnari_what_2019,zhao_diverse_nodate}. 
However, in practice it is not ideal to treat action sequences this way. Robots, for example, want to know alternative futures in order to adjust plans when contingencies and errors happen~\cite{albrecht_autonomous_2018,heard_diagnostic_2018}.  
When an alternative prediction is needed, random sampling neither ensures that all of the most likely $x$ futures are found within as many samples, nor that the first few sampled futures are the best alternatives. 
We introduce \textbf{P}redicting the \textbf{L}ikely \textbf{A}lternative \textbf{N}ext \textbf{B}est, or \textbf{PLAN-B}, which automatically finds not only the most likely predicted sequence, but likely alternative predictions as well.

Our approach consists of two parts: (i) the Collaborative RNN (Recurrent Neural Network) and (ii) the Choice Table. 
In the action prediction literature, systems are generally either trained based on per-frame features or on symbolic action sequence information \cite{Farha2018b,qi_predicting_2017}. 
These two modalities share complementary information. Action sequences provide sequence pattern information, while the features provide more fine-grained information from the frames themselves. 
Our approach combines these two methods with the Collaborative RNN. A lower level RNN captures low-resolution feature information. 
A higher level RNN captures more abstract patterns across action sequences. The two layers pass messages between one another, or ``collaborate'' in order to generate optimal predictions by utilizing information from multiple resolutions. 
This combination of features ensures that the system is able to find patterns (and thus alternative predictions) across multiple layers of resolution.

The Choice Table is a two dimensional array that contains the different predicted future actions stacked on top of one another. 
It also coordinates different decoders to give diverse future predictions. 
To accomplish ensure that the top best predictions are found, we introduce Similarity Penalty (SP) and Random Loss Negation (RLN) techniques. 
The former penalizes decoders for predicting the same action at the same time in different sequences, ensuring that the system maximizes the number of alternative sequences found. 
The latter complements this process by randomly removing standard loss from incorrect predictions, forcing the decoders to find alternative answers.

Further, current methods do not measure the ability of a system to find alternatives as such. Present metrics only measure the accuracy of the most likely action sequence, or the likelihood of the next action \cite{morais_learning_2020,zhao_diverse_nodate,piergiovanni_learning_nodate}. 
We extend accuracy measurements to include per frame mean-over-class accuracy@k.

We also introduce a new metric, Choice F1, the harmonic mean between the accuracy@k  and the average mean over class accuracy of each prediction. 
This metric extends the notion of F1 score to potential future predictions measuring the trade-off between the ability of a system to predict the most likely futures and producing a diverse set of possible futures, and gives a concise scalar representation of overall performance of a system as such. 
In total, our contributions are:
\begin{enumerate}
    \itemsep0em 
    \item We extend action prediction to also evaluate accurate, alternative predictions, introducing a new metric, Choice F1 to measure a system's ability to predict these.
    \item We present PLAN-B, a system for action prediction which automatically finds the top futures by leveraging novel Collaborative RNN and Choice Table methods.
    \item We show that our system significantly outperforms state-of-the-art systems on benchmark datasets.
\end{enumerate}

\section{Related Work}
\subsection{Action Prediction}
Work has been done in action anticipation to classify a short video before completion \cite{Kong2018a,cai_action_2019,Kong2018a}. 
More recently, there has been a boom in research for action prediction to predict futures based on seeing a percentage of a multi-action video (also called action forecasting). 
Farha et al.\  \cite{Farha2018b} proposed a CNN and RNN-based system for this problem, and were able to generate predictions from ground-truth action labels and from features derived from video frames. 
This paper predicted up to 50\% of the video after observing 20\% to 30\%. This system also started the trend of evaluating separate versions of the predictor based on ground-truth action sequences and frame features. 
Ke. et al.\ \cite{ke_time-conditioned_2019} incorporated time itself as a useful feature for prediction. 
Combined with an attention mechanism, this allowed their system to predict the future without the need for neither iterative RNN steps nor beforehand knowledge of the total length of the video. Recently, Piergiovanni et al.\ \cite{piergiovanni_learning_nodate,piergiovanni_adversarial_2020} utilized grammar structures to help guide action prediction systems. 
In these models, differentiable context free grammar models are generated in a probabilistic, adversarial manner in order to provide structure prediction sequences. 
Relatedly, hierarchical RNN systems and sequence to sequence models have also been proposed  \cite{ng_forecasting_2020,morais_learning_2020,farha_long-term_2020,sener_temporal_2020-1}.

\subsection{Probabilistic Modeling of Action Prediction}
There have been several papers that conceptualize action sequences probabilistically, and model action prediction in terms of sampling from a distribution.Farha et al.\ \cite{abu_farha_uncertainty-aware_2019} first introduced this concept by predicting future actions as samples from a distribution generated by an RNN. 
This method builds on previous RNN decoder notion, but sampled future action probabilistically from a distribution. Mehrasa et al.\ \cite{mehrasa_variational_2019} expanded on this notion with the inclusion of variational autoencoders. 
This system implemented a prior and posterior net from RNN features to sample a latent variable. From this the system was able to use a decoder to sample future actions probabilistically. Zhao and Wildes \cite{jang_categorical_2017,zhao_diverse_nodate} further developed this notion with generative adversarial methods, using Gumbel Softmax as a means of sampling future action sequences. 
This system utilized a distance regularizer in order to encourage more diverse predictions. Most recently, grammar-based system have been used to generate potential paths forward given learned structures. 
These systems combine differentiable context free grammar production with a neural network design in order to unfold action predictions over time \cite{piergiovanni_adversarial_2020,piergiovanni_learning_nodate}.

Although future actions are probabilistic in the sense that they occur with a degree of uncertainty that can be modeled with a probability distribution, there are limitations to using this paradigm with respect to the end goals of action prediction.
In scenarios such as human robot collaboration, a robot uses action prediction to try to understand future actions in order to best collaborate with the human \cite{albrecht_autonomous_2018,heard_diagnostic_2018}.
Sampling plans from a distribution provides a way to obtain likely futures, but does not specifically give the top prediction and contingencies to it.
In an attempt to sample the top five futures, for example, a system may very well sample the same future (or multiple very similar) futures multiple times.
In order to better estimate the best future and plan for likely alternatives, it would be better to simply predict the top k futures for a system.
PLAN-B provides this by finding all the top k futures without the need for wasteful sampling.

\subsection{Evaluation Metrics}
The majority of papers utilize mean over class accuracy as a protocol, assessing single predictions \cite{Farha2018b,Aakur2018,zhao_diverse_nodate}.
More recent works have taken a probabilistic approach, predicting next action log likelihood, as well as the change in accuracy as more and more future action samples are generated \cite{zhao_diverse_nodate,mehrasa_variational_2019,piergiovanni_adversarial_2020}.
These metrics, however, do not take choice or alternatives into account when predicting action sequences.
For datasets with a high level of unpredictability, such as 50 Salads, a system will only be able to become so accurate before all predictable patterns are learned \cite{morais_learning_2020}.
Per-frame mean over class accuracy cannot accommodate inherent unpredictabilities in sequences, and alternative methods do not evaluate a system's knowledge of potential contingencies or alternatives.
Our choice of metrics, accuracy@k and the novel Choice F1, do this by specifically evaluating the ability of a system to maintain accurate predictions while finding plausible different sequences.

\section{Predicting Alternatives with PLAN-B}
In order to make accurate, long-range predictions about the future, we introduce \textbf{PLAN-B}, a sequence-to-sequence model which utilizes a multilayer, ``collaborative'' recurrent neural network to encode features and decode action sequences and a Choice Table to ensure that all probable predictions are found.
A full overview of the system is shown in Figure \ref{fig:main_system}.

\subsection{Problem Definition and Data Encoding}
Formally, given a series of T frames $x_1\dots x_T$, which correspond to a list of activities $y_1\dots y_n$, we want to predict the remaining actions $y_{n+1}\dots y_N$ as well as the time lengths for these actions $\ell_{n+1}\dots \ell_N$.

We begin by extracting RGB and optical flow features from the video frames with I3D \cite{Carreira2017}.
From this, we generate per-frame features $f_1\dots f_T$, using MS-TCN++ \cite{li_ms-tcn_2020}.
These features then serve as input to the CRNN, which serves to both encode video features and then decode action sequences.
We also calculate loss on the output of the MS-TCN++ logits as action recognition loss has been shown to increase prediction accuracy \cite{farha_long-term_2020}.
\begin{align}
    \Lagr{}_\text{MS-TCN++} = \frac{1}{T}\sum_{t=1}^T -\log(\hat{y}_{t,c})
\end{align}
This is shown in step one of Figure \ref{fig:main_system}.

\subsection{Collaborative RNN}
For action prediction, most methods take as input either feature level information or action sequence and their duration \cite{Farha2018b,piergiovanni_learning_nodate,mehrasa_variational_2019}.
We introduce a novel sequence-to-sequence model for encoding observed video data and predicting future actions.
In a sequence-to-sequence model, an RNN builds an encoding over one input sequence.
The hidden state of this RNN, the encoder vector, is used as the initial hidden state for a decoder RNN, which recursively predicts future sequence tokens \cite{sutskever_sequence_2014}.

Our model leverages two GRUs (Gated Recurrent Units) operating on these different modalities; each takes the output of the other as input.
More specifically, for the encoder, the lower GRU takes as input the previous action classified by the upper RNN and the features from the MS-TCN++ layer.
At each step, the hidden state of this layer is passed through a fully connected layer in order to produce an action classification.
This action is forced into an embedding, and then used as input to the second layer.
The second layer's hidden state is, in turn, passed through a fully connected layer in order to produce another action label.
This is fed through another embedding and used as input for the next step of the lower RNN.

\begin{figure}[t]
    \centering
    \includegraphics[width=\linewidth]{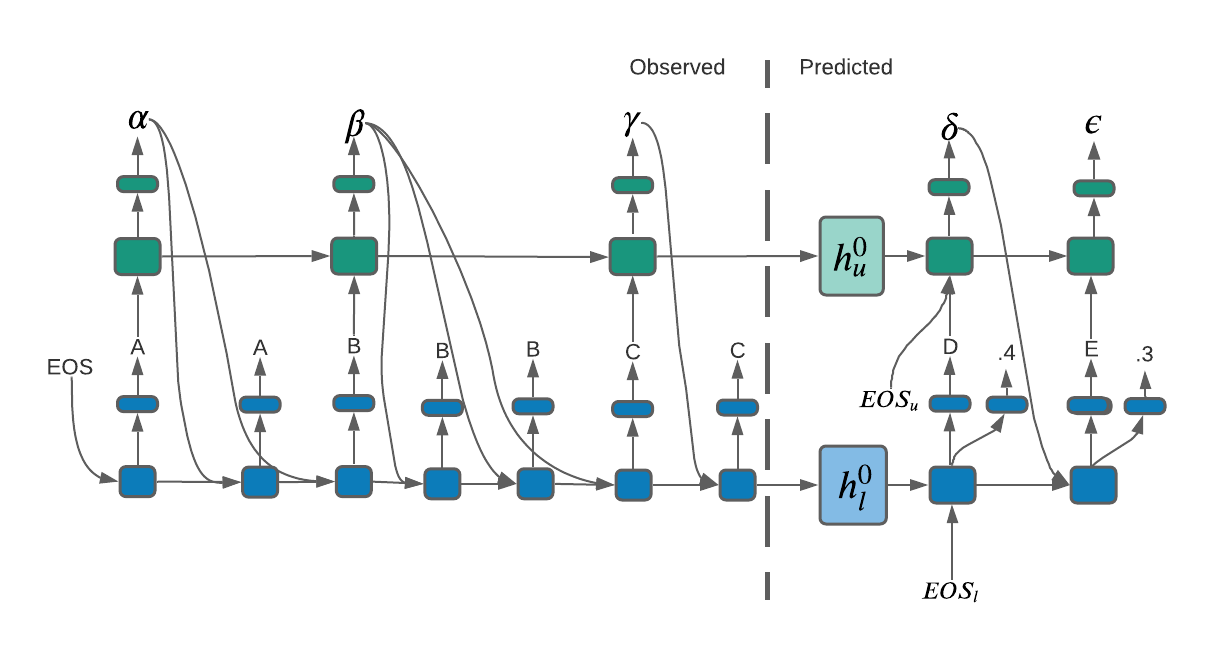}
    \caption{A detailed diagram of the CRNN. The lower RNN takes, as input, feature information. The upper level, by comparison, takes sequence information decoded by the first layer. The two layers learn patterns across separate, but essential, modalities for action prediction.}
    \label{fig:crnn_closeup}
\end{figure}

The lower RNN updates on every frame, while the upper RNN only updates when the action classification from the lower RNN changes.
In this way, the lower level learns a feature representation over the frame features, while the upper RNN learns a representation over action sequences and durations (step two of Figure \ref{fig:main_system}).

The end result of this is two separate hidden states: one from the upper, and one from the lower RNN.
These are used as inputs to the decoder RNNs.
As before, there are two GRUs per thread, one for individual level actions and one for higher level sequence information.
The lower level decodes individual actions, taking as input the previous action and the previous upper-level action. 
For both layers, the initial action for the GRUs is a special end of sequence (EOS) action.
The hidden state output of this GRU is fed into a fully connected layer to derive the action label and the relative duration for this step.
The action is fed into another embedding layer and used as input to the upper level GRU.
The upper level GRU is then fed through another fully connected layer to produce logits, the action of which is then plugged back into the lower layer RNN.
The output from the first RNN forms the end prediction sequence.
The decoder continues to predict actions until the special EOS action is predicted.
From these outputs, we calculate the cross entropy loss of the actions versus the ground truth.
This is illustrated in step three of Figure \ref{fig:main_system}.

We extract cross entropy loss from the classification outputs of the first encoder RNN layer in a manner similar to the MS-TCN++ layer
\begin{align}
    \displaystyle
    \Lagr{}_\text{RNN} = \frac{1}{T}\sum_{t=1}^T -\log(\hat{y}_{t,c})
\end{align}

Cross entropy loss is calculated for each prediction in each thread from each decoder:
\begin{align}
    \Lagr{}_\text{action} = \frac{1}{(N-n)K}\sum_{k=1}^K \sum_{m=n+1}^{N} -\log(\hat{y}_{k,m,c})
\end{align}
We utilize mean squared error as the loss in time prediction.
\begin{align}
    \Lagr{}_\text{time} = \frac{1}{(N-n)K}\sum_{k=1}^K \sum_{m=n+1}^{N} (\hat{\ell}_{k,m} - \ell_{m})^2
\end{align}
Where $\hat{y}_{k,m,c}$ is the probability of predicting the ground truth class c at point m in thread k.
$\hat{\ell}_{k,m}$ is the prediction for the predicted duration of the activity at position $m$ in thread k after passing all estimations for time for that action sequence $\ell_{k, \cdot}$ through a softmax function.
A detailed visualization of the CRNN is shown in Figure \ref{fig:crnn_closeup}.

\subsection{Choice Table}
PLAN-B uses $k$ decoders to make $k$ separate predictions.
The Choice Table is a $\mathbb{R}^{K\times (N-n)}$ table representing the total symbolic action predictions for each thread $k \leq K$ and each spot in the predicted sequence $(n+1) \leq m \leq N$.
Due to the long range of forecasting and the inherent unpredictability of the problem, a decoder can occupy one of several local minima in weight space which produce similarly likely predictions.
For example, given the input sequence [A,B], the actions C or D might occur with equal likelihood.
Only one can be technically correct for any given test sequence, but this is not necessarily predictable.
A decoder will either learn to predict one or the other.
A decoder that predicts each occupies a local minimum with a specific set of learned weights.
We find this phenomenon validated as different training runs with different initial weights produced predictors that made different predictions, but nonetheless have similar accuracy.

Because each minimum is similarly optimal, it would be best to create a system that for finding all minima.
In order to accomplish this, we utilize two novel mechanisms- Similarity Penalty and Random Loss Negation.
The Choice Table is shown in step 4 of Figure \ref{fig:main_system}.

\subsubsection{Similarity Penalty}
We observe that having separate encoders with separate, randomly initialized weights is insufficient to prevent the threads from converging to the same minima.
In order to push the threads into different local minima, we add a Similarity Penalty based on the $L2$ distance between the output softmax logits in the system.
\begin{align}
    \Lagr{}_\text{sim} = -\frac{1}{K^2}\lambda \vec{e}^TD(\textbf{Q},\textbf{Q})\vec{e} \label{eq:lambda}
\end{align}

Where $\vec{e}$ is a column vector whose entries are all ones, $\textbf{Q} \in \mathbb{R}^{K \times (N-n) \times C}$ is the matrix representing the softmax logit outputs for all $K$ threads, each having at most $N-n$ actions over $C$ classes, and  $D(\cdot, \cdot)$ computes the matrix of pairwise distances between elements in all threads.

\subsubsection{Random Loss Negation}
\begin{figure}
    \centering
    \includegraphics[scale=.3]{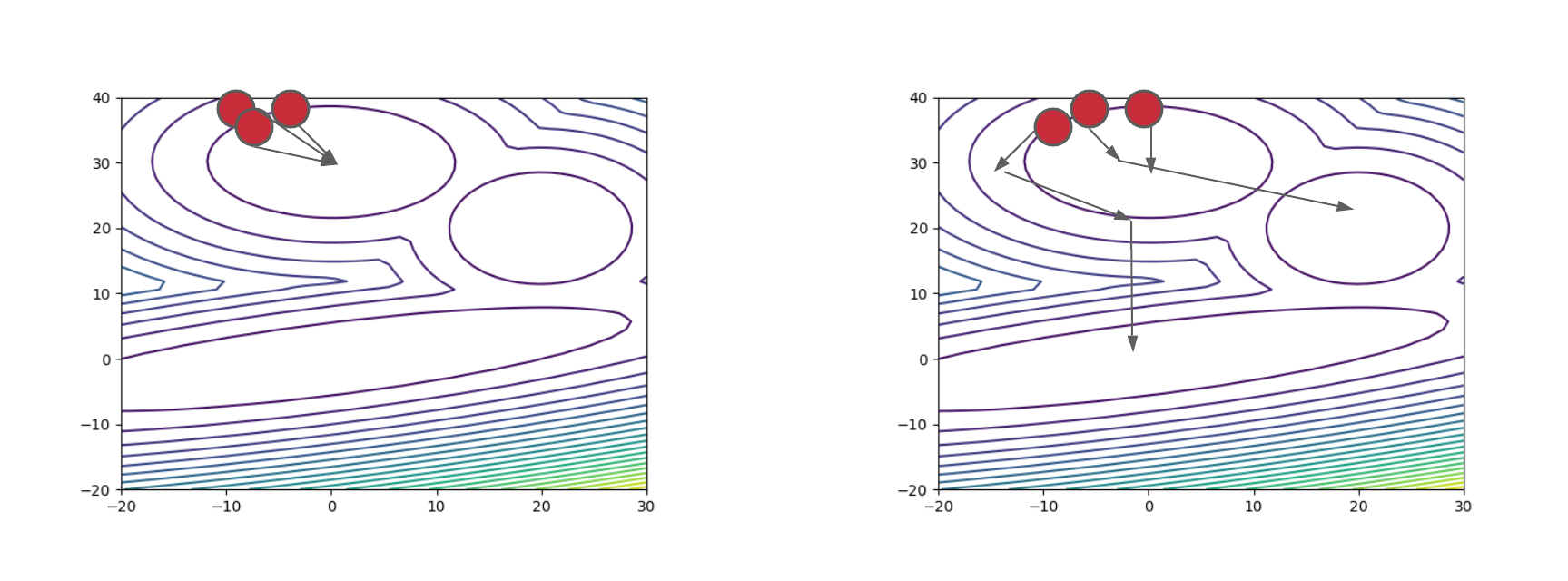}
    \caption{Illustration of how the Similarity Penalty and Random Loss Negation affect gradient descent in the decoder weights. Each decoder weight (red) will flow to the same local minimum (left). With Similarity Penalty and Random Loss Negation, the gradient is perturbed so that the decoders are repelled in weight space (right). This ensures that all local minima are found.}
    \label{fig:dropout_illustrated}
\end{figure}

The Similarity Penalty is, however, insufficient to ensure diversity of outputs.
Although the similarity incentivizes the system to find different weights, it is ultimately overpowered by the loss function.
However, if the weights are allowed to take steps away from each other (based on similarity), with no interference from the prediction loss, then they will predict different probable threads. 
A full illustration of this process is shown in Figure \ref{fig:dropout_illustrated}.
In order to create this process, we create a mask $M \in \mathbb{R}^{K \times N}$ over the Choice Table.
This randomly generated mask's elements are drawn from a Bernoulli distribution, with a probability $\phi$.
Elements with a 1 have their loss values set to zero.
\begin{align}
    \forall i \in [1\dots K], j \in [1\dots N] M_{ij} \sim B(1,\phi) \label{eq:phi}
\end{align}

By randomly canceling out the loss penalty at each step, gradient descent seeks instead to maximize the distance between decoder weights.
This allows them to each enter different local minima, and thus predict different likely sequences.
Note that this is not sampling future actions as proposed by previous works, but, instead, using sampling as a training technique to optimize our system.
The result of both the Similarity Penalty and the Random Loss Negation is a system in which decoders are not only drawn to minima, but repelled frequently from one another.
The action loss thus becomes:
\begin{align}
    \Lagr{}_\text{action} = \frac{1}{(N-n)K}\sum_{k=1}^K \sum_{m=n+1}^{N} -M[k,m]\log(\hat{y}_{k,m,c})
\end{align}
The overall loss function for the system is thus
\begin{align}
    \Lagr{}_\text{total} = \Lagr{}_\text{MS-TCN++} + \Lagr{}_\text{RNN} + \Lagr{}_\text{action} + \Lagr{}_\text{time} + \Lagr{}_\text{sim}
\end{align}

\begingroup
\renewcommand{\arraystretch}{.9}
\begin{table*}
    \centering
    \begin{tabular}{|c|c||c c c c | c c c c|}
        \hline
        \textbf{Dataset} & \textbf{Observation} &  \multicolumn{4}{c|}{20\%} & \multicolumn{4}{c|}{30\%}\\\hline\hline
         & \textbf{Prediction}  & 10\% & 20\% & 30\% & 50\% & 10\% & 20\% & 30\% & 50\% \\\hline
        \multirow{3}{*}{50 Salads} & Nearest Neighbor \cite{Farha2018b} & 19.04 & 16.10 & 14.13 & 10.37 & 21.63 & 15.48 & 13.47 & 13.90 \\
        & RNN \cite{Farha2018b} & 30.06 & 25.43 & 18.74 & 13.49 & 30.77 & 17.19 & 14.79 & 9.77 \\
        & CNN \cite{Farha2018b} & 21.24 & 19.03 & 15.98 & 9.87 & 29.14 & 20.14 & 17.46 & 10.86 \\
        & Uncertainty Aware \cite{abu_farha_uncertainty-aware_2019} & 24.86 & 22.37 & 19.88 & 12.82 & 29.10 & 20.50 & 15.28 & 12.31 \\
        & Time Conditioned \cite{ke_time-conditioned_2019} & 32.5 & 27.60 & 21.3 & 16.0 & 35.10 & 27.10 & 22.1 & 15.6 \\
        & Cycle Consistency \cite{farha_long-term_2020} & 34.76 & 28.41 & 21.82 & 15.25 & 34.39 & 23.70 & 18.95 & 15.89 \\
        & Adversarial Grammar \cite{piergiovanni_adversarial_2020} & \bf{39.5} & \bf{33.2} & 25.9 & 21.2 & 39.5 & \bf{31.5} & 26.4 & 19.8 \\
        & \textbf{PLAN-B} (OURS) & 37.58 & 31.12 & \textbf{26.65} & \textbf{21.93} & \textbf{40.12} & 30.92 & \textbf{26.44} & \textbf{20.98} \\\hline
        \multirow{3}{*}{Breakfast} & Grammar \cite{Farha2018b} & 16.60 & 14.95 & 13.47 & 13.42 & 21.10 & 18.18 & 17.46 & 16.30 \\
        & Nearest Neighbor \cite{Farha2018b} & 16.42 & 15.01 & 14.47 & 13.29 & 19.88 & 18.64 & 17.97 & 16.57 \\
        & Uncertainty Aware \cite{abu_farha_uncertainty-aware_2019} & 16.71 & 15.40 & 14.47 & 14.20 & 20.73 & 18.27 & 18.42 & 16.86 \\
        & CNN \cite{Farha2018b} & 17.90 & 16.35 & 15.37 & 14.54 & 22.44 & 20.12 & 19.69 & 18.76 \\
        & RNN \cite{Farha2018b} & 18.11 & 17.21 & 16.42 & 15.84 & 22.75 & 20.44 & 19.64 & 19.75 \\
        & Time Conditioned \cite{ke_time-conditioned_2019} & 18.41 & 17.21 & 16.42 & 15.84 & 22.75 & 20.44 & 19.64 & 19.75 \\
        & Cycle Consistency \cite{farha_long-term_2020} & 25.88 & 23.42 & 22.42 & 21.54 & 29.66 & 27.37 & 25.58 & 25.20 \\
        & \textbf{PLAN-B} (OURS) & \textbf{33.49} & \textbf{32.02} & \textbf{30.75} & \textbf{29.52} & \textbf{40.57} & \textbf{37.89} & \textbf{36.06} & \textbf{34.91}\\\hline
    \end{tabular}
    \caption{Accuracy@1 results on the 50 Salads and Breakfast datasets.}
    \label{tab:top1}
\end{table*}
\endgroup

At inference time, after the predictions are made, the probability for each predicted sequence is evaluated by multiplying the logit probabilities of each output class:
\begin{align}
    \displaystyle
    P_k = \sum_{m=n+1}^N \log(\hat{y}_{k,m,c})
\end{align}

We then sort by this value, ensuring that the top prediction is the most probably prediction, the second prediction is the second most probable etc.

\section{Experiments and Results}
We begin by evaluating PLAN-B on two benchmark datasets using mean over class (MoC) accuracy@1.
We use the standard protocol of observing a small percentage of the video and then predicting the remaining 50\% \cite{Farha2018b}.
In addition, we demonstrate PLAN-B's ability to find all likely alternate predictions with accuracy@k and our proposed metric, Choice F1 using the same percent observed/percent predicted paradigm.

\subsection{Datasets}
We evaluate our model on Breakfast and 50 Salads, two leading benchmarks for action prediction \cite{Kuehne2014a,stein_combining_nodate}.
The Breakfast dataset contains 712 videos, each several minutes in length and shot with multiple camera angles. 
Each video belongs to one out of ten activities, such as making coffee or scrambling eggs.
The frames are annotated with fine-grained labels like ``pour cream'' and ``grab cup''. 
There are 48 actions total.
For evaluation, we use the standard 4 splits as proposed, each with 252 videos used for test and the rest used for training.

The 50 Salads dataset contains 50 videos showing people preparing salads. 
The average duration is 6.4 minutes, and 20 actions per video.
In addition to start and end action labels, there are 17 fine-grained action labels such as ``cut cucumber'' and ``add oil.'' 
For evaluation, we use five-fold cross-validation and report the average where ten videos are held out for test in each split.

\subsection{Implementation Details}
For our network, we set the $\lambda$ value from Equation \ref{eq:lambda} to 0.1 and the $\phi$ value from Equation \ref{eq:phi} value to 0.9.
Our system is implemented using PyTorch.
We train using established protocols in the area by allowing the network to observe the first 20\%-30\% of the video and predict the remaining 50\% \cite{farha_long-term_2020,Farha2018b}.
We extract I3D features from the videos with a network pretrained on the Kinetics dataset \cite{Kay2017b,Carreira2017}.
We train for 80 epochs with an Adam optimizer with an initial learning rate of 0.001.
Our learning rate is multiplied by 0.8 every 20 epochs.
For each experiment, we train each of the k folds three times for convergence. 
We sample frame data down to 5 frames per second before inputting it into the system.
Unless otherwise specified, we use 10 threads for our model.

\subsection{Baselines}
We first present our system's results using standard MoC accuracy.
This corresponds to accuracy@k for k=1.
We compare against state-of-the-art systems: the RNN and CNN model from Farha et al.\  \cite{Farha2018b};
their Cycle Consistency model \cite{farha_long-term_2020}; 
their Uncertainty Aware system \cite{abu_farha_uncertainty-aware_2019},
the Time Conditioned system from Ke et al.\ 2019 \cite{ke_time-conditioned_2019}; and The Adversarial Grammar System by Piergiovanni et al.\  \cite{piergiovanni_adversarial_2020}.

\subsection{Accuracy@1 Experiments}
The results for the 50 Salads ad Breakfast dataset are shown in Table \ref{tab:top1}.
Our system achieves top performance in most measurements for the 50 Salads dataset.
Our results on Breakfast even more strongly show the improvements through our system.
The greatest improvements for PLAN-B occur after observing more of the video, and for making longer-term forecasts.
This indicates that our system is particularly good at predicting longer-term futures, especially when more data is observed.

\subsection{Accuracy@k Experiments}
We also demonstrate the efficacy of our network with the accuracy@k metric for k=3.
This demonstrates the ability of a system to not only find the best prediction, but also the top two alternatives.
For this we compare our system against two state-of-the-art systems \cite{farha_long-term_2020,Farha2018b}.
Our results for 50 Salads and Breakfast datasets can be seen in Table \ref{tab:top3}.
As before, our system is able to outperform existing baselines.
In particular, the further out predictions are in the future and the more observation there is of the video, the better PLAN-B performs.
To obtain the $k$ separate choices for the accuracy@k baselines, we retrain it $k$ times with different random seeds to ensure that no two models converge to similar results.

\begingroup
\renewcommand{\arraystretch}{.9}
\begin{table*}[]
    \centering
    \begin{tabular}{|c|c||c c c c | c c c c|}
        \hline
        \textbf{Dataset} & \textbf{Observation} &  \multicolumn{4}{c|}{20\%} & \multicolumn{4}{c|}{30\%}\\\hline\hline
        & \textbf{Prediction} & 10\% & 20\% & 30\% & 50\% & 10\% & 20\% & 30\% & 50\% \\\hline
        \multirow{3}{*}{50 Salads} & RNN \cite{Farha2018b} & 31.89 & 23.13 & 20.18 & 15.13 & 34.51 & 21.4 & 16.55 & 9.96 \\
        & Cycle Consistency \cite{farha_long-term_2020} & \textbf{56.96} & 45.92 & 36.69 & 30.87 & 44.96 & 36.10 & 33.39 & 27.45 \\
        & \textbf{PLAN-B} (OURS) & 52.77 & \textbf{46.42} & \textbf{42.27} & \textbf{38.44} & \textbf{52.59} & \textbf{47.25} & \textbf{42.26} & \textbf{39.62} \\\hline
        \multirow{3}{*}{Breakfast} & RNN \cite{Farha2018b} & 22.49 & 21.38 & 21.12 & 21.92 & 27.51 & 27.19 & 27.36 & 28.74 \\
        & Cycle Consistency \cite{farha_long-term_2020} & 32.22 & 30.18 & 29.41 & 27.83 & 34.9 & 33.36 & 31.00 & 30.15 \\
        & \textbf{PLAN-B} (OURS) & \textbf{38.55} & \textbf{38.21} & \textbf{38.09} & \textbf{37.72} & \textbf{44.89} & \textbf{43.5} & \textbf{42.86} & \textbf{42.42} \\ \hline
    \end{tabular}
    \caption{Accuracy@3 results on the 50 Salads and Breakfast datasets.}
    \label{tab:top3}
\end{table*}
\endgroup

\subsection{Ablation Study}
We also analyze the impact of individual components of our system in order to determine the benefits of each.
The results for 50 Salads can be seen in Table \ref{tab:ablation_50s}.
The results indicate that each component provides significant gains in accuracy, with the inclusion of the SP and RLN providing the largest marginal gain.
Interestingly, the Collaborative RNN provides a higher gain than selecting from ten threads (up from three).
The CRNN in particular increases the accuracy of predictions where more of the video is observed.

\begin{table}[]
    \centering
    \begin{tabular}{|c | c | c|}
    \hline
        Observation \% &  20\% & 30\% \\\hline
        Prediction \% & 50\% & 50\% \\\hline
        Multiple decoder system & 17.76 & 18.43 \\
        + SP, RLN & 37.09 & 33.53 \\
        + Selecting from 10 threads & 37.69 & 36.19 \\
        + Collaborative RNN & 38.44 & 39.62 \\ \hline
    \end{tabular}
    \caption{Ablation study on 50 Salads.}
    \label{tab:ablation_50s}
\end{table}

\subsection{Accuracy vs. Total Number of Threads}
The more alternative futures one selects from, the more likely it is that one will find improvements on predictability accuracy.
In order to investigate this hypothesis, we vary the number of  prediction threads for our system and note the change in accuracy. 
The results on the 50 Salads dataset can be seen in Table \ref{tab:threads50s}.
The results indicate that increased number of threads leads to increased accuracy.
However, there are diminishing returns for each additional thread.
More than doubling the number of threads from 20 to 50 only marginally increases the end accuracy.
 
\begin{table}[]
    \centering
    \begin{tabular}{|c | c | c|}
    \hline
        Observation \% &  20\% & 30\% \\\hline
        Prediction \% & 50\% & 50\% \\\hline
        5 threads & 38.13 & 38.18 \\
        10 threads & 38.44 & 39.62 \\
        20 threads & 40.05 & 39.93 \\
        50 threads & 42.13 & 40.14 \\\hline
    \end{tabular}
    \caption{Accuracy vs. number of Threads on 50 Salads.}
    \label{tab:threads50s}
\end{table}

\subsection{Choice F1 for Evaluating Trade-offs Between Alternatives and Accuracy}
\begin{figure*}
\begin{subfigure}{.5\textwidth}
  \centering
  \includegraphics[scale=0.5]{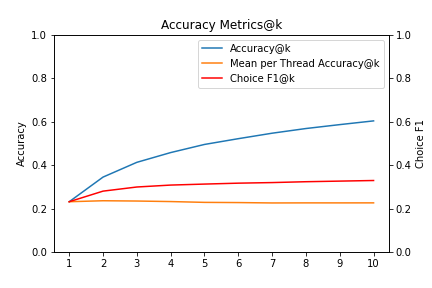}
  \label{fig:sub-first}
\end{subfigure}
\begin{subfigure}{.5\textwidth}
  \centering
 \includegraphics[scale=0.5]{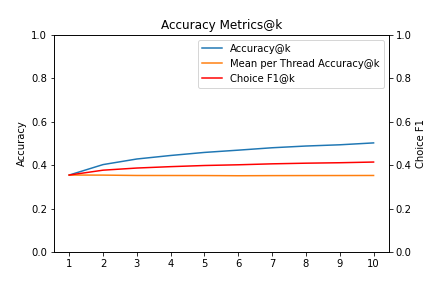}
  \label{fig:sub-second}
\end{subfigure}
\caption{The results of the F1 experiment on the 50 Salads dataset (left) and Breakfast (right). The F1 Choice metric demonstrates that our network is able to predict all the most likely threads for a given video.}
\label{fig:choicef1}
\end{figure*}

Although accuracy@k provides a more robust understanding of the system as a whole, it is not an ideal metric.
As $k$ goes to infinity, the total number of different actions increases dramatically.
After a certain point $k$ it becomes possible for the ``ideal'' system to be no better than one which randomly predicts actions.
There is thus a need to measure the trade-off between the accuracy for each prediction and ensuring that all plausible choices are covered.
To measure this, we introduce a novel metric- Choice F1.
This metric, based on the F1 score, is the harmonic mean between mean per-thread accuracy and accuracy@k:
\begin{align}
    \displaystyle
    F1_c = \frac{2(MPTA@k*Acc@k)}{(MPT@k + Acc@k)}
\end{align}
where $MPT@k$ is the mean per thread accuracy @k and $Acc@k$ is the accuracy@k.
These metrics on the 50 Salads and Breakfast dataset with increasing k can be seen in Figure \ref{fig:choicef1}.
Our results show that Choice F1 highlights the fact that as k increases, the accuracy@k increases, but the per-thread accuracy does not change.
There are also diminishing returns as k increases.
The metric also demonstrates that the 50 Salads dataset is more under-determined than Breakfast.
This is to be expected, as there are more plausible alternative sequences in 50 Salads than in Breakfast.

\begin{figure*}
\centering
\begin{subfigure}[b]{1\linewidth}
\includegraphics[width=\linewidth, height=6em]{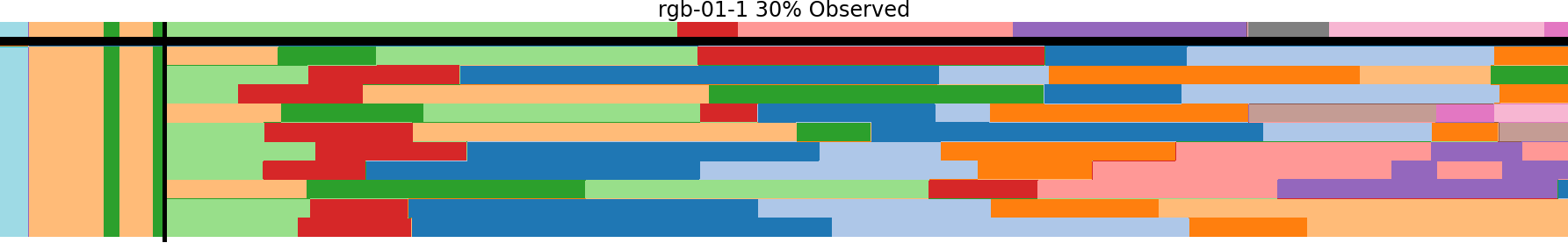}
\end{subfigure}

\begin{subfigure}[b]{1\linewidth}
\includegraphics[width=\linewidth, height=6em]{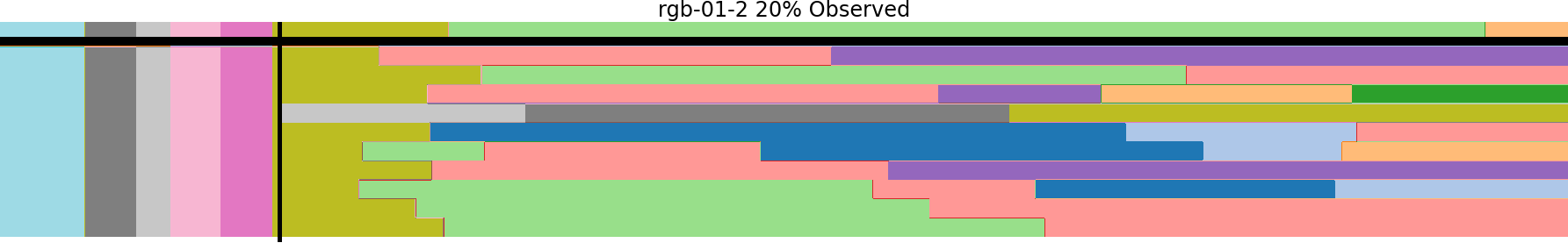}
\end{subfigure}

\begin{subfigure}[b]{1\linewidth}
\includegraphics[width=\linewidth, height=6em]{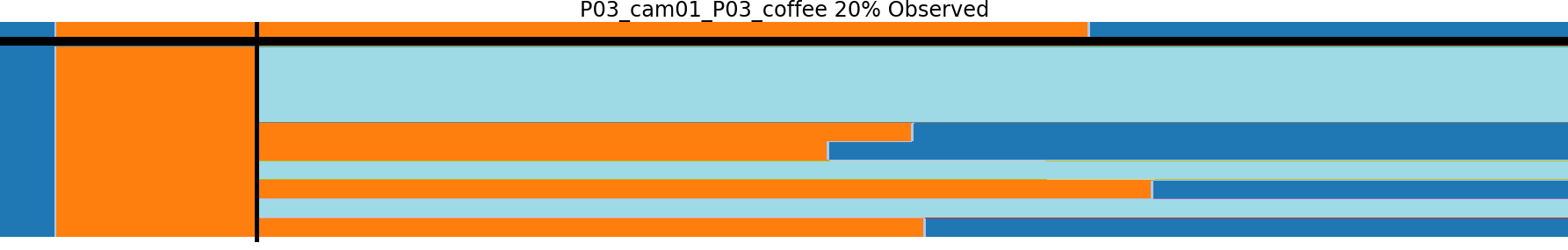}
\end{subfigure}

\begin{subfigure}[b]{1\linewidth}
\includegraphics[width=\linewidth, height=6em]{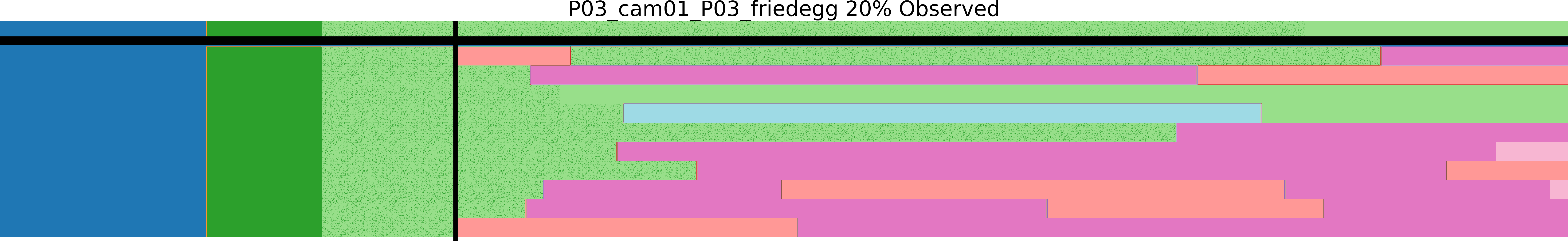}
\end{subfigure}
\caption{Qualitative results of the PLAN-B Model. The top row, separated by the horizontal black line, is a colored segmentation of the ground truth data. Subsequent rows are the system's predictions. The vertical bar represents the division between observed and predicted data.}
\label{fig:qualitative}
\end{figure*}

\section{Summary}
We demonstrate a system for anticipating not only sequences of actions, but alternatives to those as well.
This key feature allows for our system to find all probable alternative predictions in order of likelihood without the need for wasteful or impractical sampling. We show how our Choice Table  allows the system to find the top predictions, in order. The multilayer Collaborative RNN allows the system to capture multi-level action patterns, further augmenting prediction capabilities. We demonstrate the efficacy of our system by exceeding state-of-the-art results on competitive datasets. We also introduce the Choice F1 metric for evaluating the ability of a system to simultaneously find all potential futures without degenerating into random guessing. With this work, we not only demonstrate an improvement on existing systems, but we also introduce a new  perspective on action recognition.



{\small
\bibliographystyle{ieee_fullname}
\bibliography{refs}
}

\end{document}